\documentclass{article}

\usepackage[final,nonatbib]{nips_2017}
\usepackage[numbers,square,comma,sort&compress]{natbib}

\usepackage[utf8]{inputenc} % allow utf-8 input
\usepackage[T1]{fontenc}    % use 8-bit T1 fonts
\usepackage{hyperref}       % hyperlinks
\usepackage{url}            % simple URL typesetting
\usepackage{booktabs}       % professional-quality tables
\usepackage{amsfonts}       % blackboard math symbols
\usepackage{nicefrac}       % compact symbols for 1/2, etc.
\usepackage{microtype}      % microtypography
\usepackage{enumitem}

\usepackage{graphicx}
\usepackage[font=small]{caption}
\usepackage{subcaption}
\graphicspath{{figures/pdf/}}
\DeclareGraphicsExtensions{.pdf}

\usepackage{tikz}
\usetikzlibrary{bayesnet}

\usepackage{amsfonts}
\usepackage{amsmath}
\usepackage{amsthm}

\usepackage{algorithm}
\usepackage{algorithmic}

\usepackage{booktabs}
\usepackage{multirow}
\usepackage{tabularx}
\usepackage{wrapfig}

\usepackage{macro}
\usepackage{comment}
\usepackage{xcolor}

\DeclareCaptionLabelFormat{andtable}{\tablename~\thetable~\&~#1~#2}

\title{The Intriguing Properties of Model Explanations}

\author{%
  Maruan~Al-Shedivat \\
  Carnegie Mellon University \\
  \texttt{alshedivat@cs.cmu.edu} \\
  \And
  Avinava~Dubey \\
  Carnegie Mellon University \\
  \texttt{akdubey@cs.cmu.edu} \\
  \And
  Eric~P.~Xing \\
  Carnegie Mellon University \\
  \texttt{epxing@cs.cmu.edu}%
}

\begin{document}

\maketitle

\begin{abstract}
Linear approximations to the decision boundary of a complex model have become one of the most popular tools for interpreting predictions.
In this paper, we study such linear explanations produced either \emph{post-hoc} by a few recent methods or generated along with predictions with \emph{contextual explanation networks} (CENs).
We focus on two questions:
(i) whether linear explanations are always consistent or can be misleading, and
(ii) when integrated into the prediction process, whether and how explanations affect performance of the model.
Our analysis sheds more light on certain properties of explanations produced by different methods and suggests that learning models that explain and predict jointly is often advantageous.
\end{abstract}

%!TEX root = ../paper.tex

\section{Introduction}

Model interpretability is a long-standing problem in machine learning that has become quite acute with the accelerating pace of widespread adoption of complex predictive algorithms.
There are multiple approaches to interpreting models and their predictions ranging from a variety of visualization techniques~\citep{simonyan2013deep,yosinski2015understanding,mahendran2015understanding} to explanations by example~\citep{caruana1999case,kim2014bayesian}.
The approach that we consider in this paper thinks of explanations as models themselves that approximate the decision boundary of the original predictor but belong to a class that is significantly simpler (e.g., local linear approximations).

Explanations can be generated either \emph{post-hoc} or alongside predictions.
A popular method, called LIME~\citep{ribeiro2016trust}, takes the first approach and attempts to explain predictions of an arbitrary model by searching for linear local approximations of the decision boundary.
On the other hand, recently proposed contextual explanation networks (CENs)~\citep{alshedivat2017cen} incorporate a similar mechanism directly into deep neural networks of arbitrary architecture and learn to predict and to explain jointly.
Here, we focus on analyzing a few properties of the explanations generated by LIME, its variations, and CEN.
In particular, we seek answers to the following questions:
\begin{enumerate}[leftmargin=2em,itemsep=-0.5pt,topsep=-1pt]
    \item Explanations are as good as the features they use to explain predictions.
    We ask whether and how feature selection and feature noise affect consistency of explanations.
    \item When explanation is a part of the learning and prediction process, how does that affect performance of the predictive model?
    \item Finally, what kind of insight we can gain by visualizing and inspecting explanations?
\end{enumerate}

%!TEX root = ../paper.tex

\section{Methods}\label{sec:methods}

We start with a brief overview of the methods compared in this paper: LIME~\citep{ribeiro2016trust}
% , its version with Shapley kernel~\citep{lundberg2017unified},
and CENs~\citep{alshedivat2017cen}.
Given a dataset of inputs, $\xv \in \Xc$, and targets, $y \in \Yc$, our goal is to learn a predictive model, $f : \Xc \mapsto \Yc$.
To explain each prediction, we have access to another set of features, $\zv \in \Zc$, and construct explanations, $g_{\xv} : \Zc \mapsto \Yc$, such that they are consistent with the original model, $g_{\xv}(\zv) = f(\xv)$.
These additional features, $\zv$, are assumed to be more interpretable than $\xv$, and are called \emph{interpretable representation} in~\cite{ribeiro2016trust} and \emph{attributes} in~\citep{alshedivat2017cen}.
% , and \emph{simplified inputs} in~\citep{lundberg2017unified}.

\subsection{LIME and Variations}\label{sec:lime}

Given a trained model, $f$, and an instance with features $(\xv, \zv)$, LIME constructs an explanation, $g_{\xv}$, as follows:
\begin{equation}
    \label{eq:LIME-general}
    g_{\xv} = \argmin_{g \in \Gc} \Lc(f, g, \pi_{\xv}) + \Omega(g)
\end{equation}
where $\Lc(f, g, \pi_{\xv})$ is the loss that measures how well $g$ approximates $f$ in the neighborhood defined by the similarity kernel, $\pi_{\xv} : \Zc \mapsto \Rb_+$, in the space of additional features, $\Zc$, and $\Omega(g)$ is the penalty on the complexity of explanation.
Now more specifically, \citet{ribeiro2016trust} assume that $\Gc$ is the class of linear models:
\begin{equation}
    \label{eq:explanation-class}
    g_{\xv}(\zv) := b_{\xv} + \wv_{\xv} \cdot \zv
\end{equation}
and define the loss and the similarity kernel as follows:
\begin{equation}
    \label{eq:LIME-specific}
    \Lc(f, g, \pi_{\xv}) := \sum_{\zv^\prime \in \Zc} \pi_{\xv}(\zv^\prime)\left(f(\xv^\prime) - g(\zv^\prime)\right)^2, \quad
    \pi_{\xv}(\zv^\prime) := \exp\left\{-D(\zv, \zv^\prime)^2 / \sigma^2\right\}
\end{equation}
where the data instance is represented by $(\xv, \zv)$, $\zv^\prime$ and the corresponding $\xv^\prime$ are the perturbed features, $D(\zv, \zv^\prime)$ is some distance function, and $\sigma$ is the scale parameter of the kernel. $\Omega(g)$ is further chosen to favor sparsity of explanations.
% \citet{lundberg2017unified} argue that when $\zv \in \Zc \equiv \{0, 1\}^M$ are binary vectors of indicators over the (subsets of) dimensions of $\xv$, the solution to \eqref{eq:LIME-general} is likely to satisfy certain consistency properties when $\Omega(g) \equiv 0$ and the so called \emph{Shapley kernel} is used:
% \begin{equation}
%     \label{eq:shapley-kernel}
%     \pi_{\xv}(\zv) := (M - 1) \bigg/ \binom{M}{|\zv|} |\zv| (M - |\zv|)
% \end{equation}
% where $|\zv| := \mathbf{1}^\top \zv$ and $M$ is the dimensionality of $\Zc$.

\subsection{Contextual Explanation Networks}\label{sec:CEN}

LIME is a \emph{post-hoc} model explanation method.
This means that it justifies model predictions by producing explanations which, while locally correct, are never used to make the predictions in the first place.
Contrary to that, CENs use explanations as the integral part of the learning process and make predictions by \emph{applying} generated explanations.
Now more formally, CENs construct the predictive model $f : \Xc \times \Zc \mapsto \Yc$ via a composition: given $\xv$, an encoder, $e_{\theta} : \Xc \mapsto \Gc$, produces an explanation $g$ which is further applied to $\zv$ to make a prediction.
In other words:
\begin{equation}
    \label{eq:CEN-simple}
    f(\xv, \zv) := g_{\xv} (\zv),\, \text{where}\ g_{\xv} := e_{\theta}(\xv)
\end{equation}
In~\citep{alshedivat2017cen} we introduced a more general probabilistic framework that allows to combine different deterministic and probabilistic encoders with explanations represented by arbitrary graphical models.
To keep our discussion simple and concrete, here we assume that explanations take the same linear form \eqref{eq:explanation-class} as for LIME and the encoder maps $\xv$ to $(b_{\xv}, \wv_{\xv})$ as follows:
\begin{equation}
    \label{eq:CEN-encoder}
    b_{\xv} := \alphav_{\theta}(\xv)^\top B, \quad
    \wv_{\xv} := \alphav_{\theta}(\xv)^\top W, \quad
    \text{where}\ \sum_{k=1}^K \alphav_{\theta}^{(k)}(\xv) = 1, \forall k : \alphav_{\theta}^{(k)}(\xv) \geq 0
\end{equation}
In other words, explanation $(b_{\xv}, \wv_{\xv})$ is constrained to be a convex combination of $K$ components from a global learnable dictionary, $D := (B, W)$, where the combination weights, $\alphav_{\theta}(\xv)$, also called \emph{attention}, are produced by a deep network.
Encoder of such form is called \emph{constrained deterministic map} in~\citep{alshedivat2017cen} and the model is trained jointly w.r.t. $(\theta, B, W)$ to minimize the prediction error.

%!TEX root = ../paper.tex

\section{Analysis}\label{sec:analysis}

Both LIME and CEN produce explanations in the form of linear models that can be further used for prediction diagnostics.
Our goal is to understand how different conditions affect explanations generated by both methods, see whether this may lead to erroneous conclusions, and finally understand how jointly learning to predict and to explain affects performance.

We use the following 3 tasks in our analysis: MNIST image classification\footnote{\url{http://yann.lecun.com/exdb/mnist/}}, sentiment classification of the IMDB reviews~\citep{maas2011learning}, and poverty prediction for households in Uganda from satellite imagery and survey data~\citep{jean2016combining}.
The details of the setup are omitted in the interest of space but can be found in~\citep{alshedivat2017cen}, as we follow exactly the same setup.

\subsection{Consistency of Explanations}\label{sec:consistency}

Linear explanation assign weights to the interpretable features, $\zv$, and hence strongly depend their quality and the way we select them.
We consider two cases where (a) the features are corrupted with additive noise, and (b) selected features are incomplete.
For analysis, we use MNIST and IMDB data.

%!TEX root = ../paper.tex

\begin{figure}[t]
    \centering
    \begin{subfigure}[b]{0.45\textwidth}
        \includegraphics[width=\textwidth]{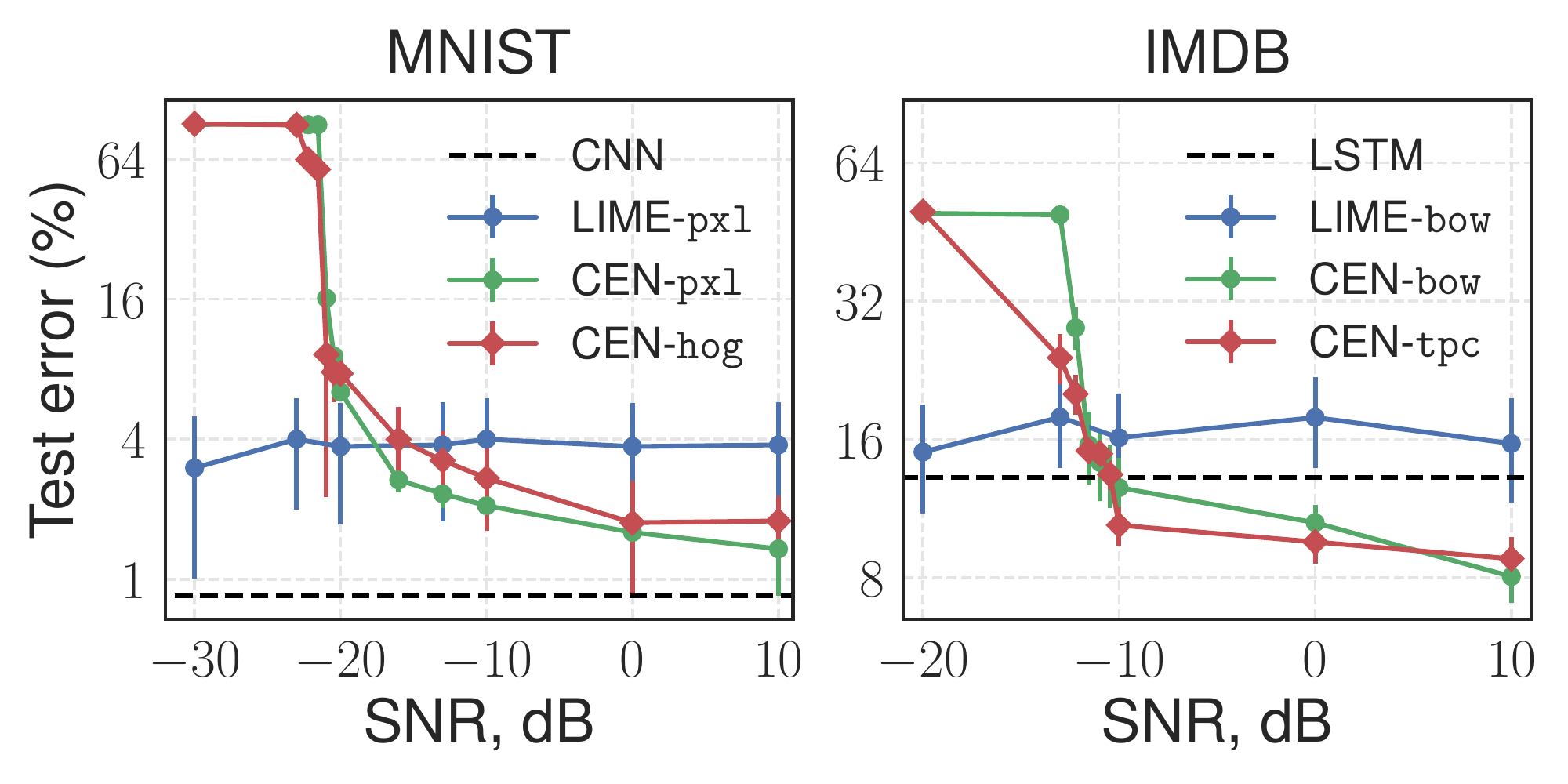}%
        \vspace{-2ex}\caption{}\label{fig:signal-to-noise}
    \end{subfigure}
    \quad
    \begin{subfigure}[b]{0.45\textwidth}
        \includegraphics[width=\textwidth]{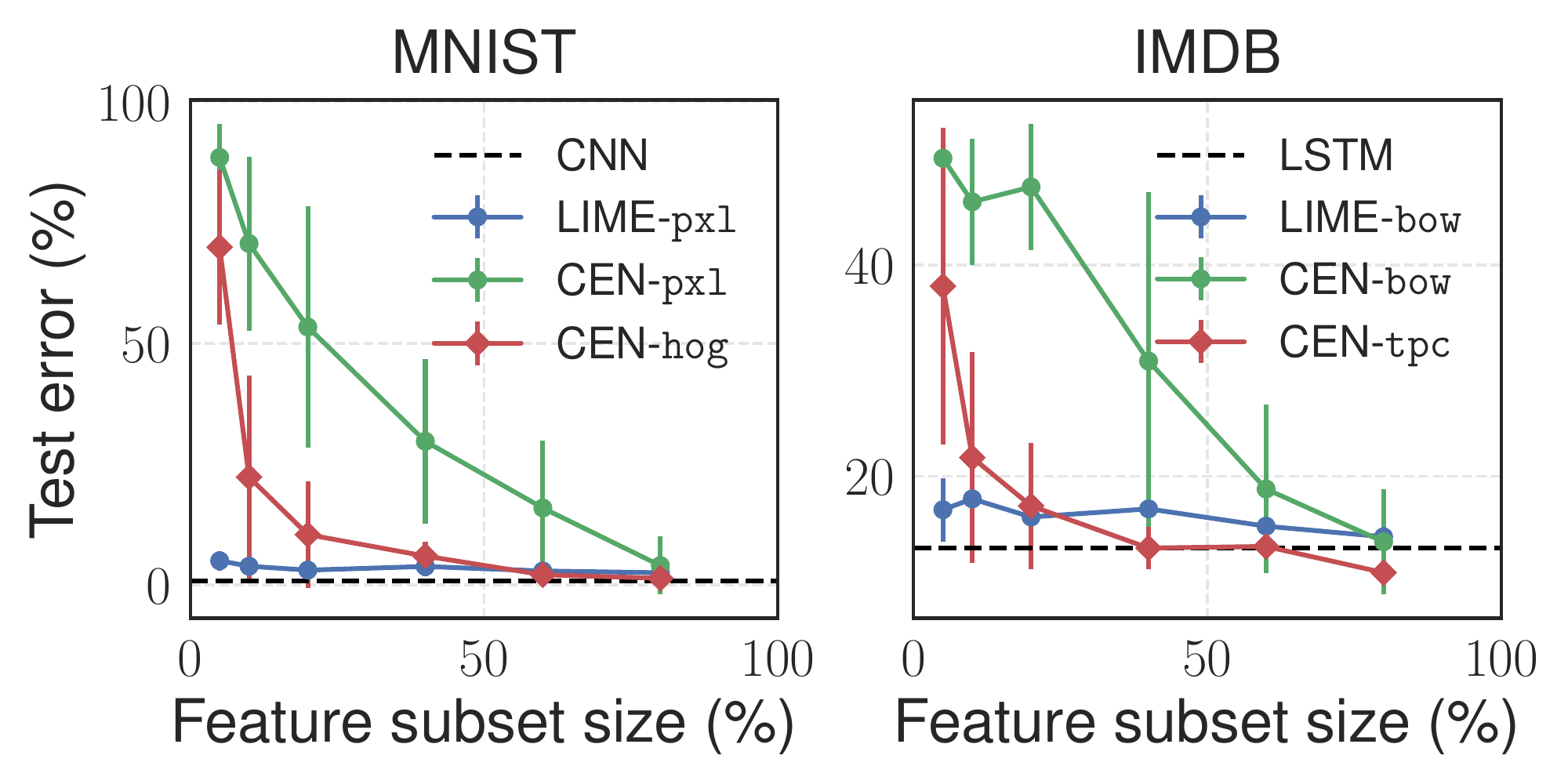}%
        \vspace{-2ex}\caption{}\label{fig:incomplete-features}
    \end{subfigure}%
    \caption{%
    The effect of feature quality on explanations.
    (a) Explanation test error vs. the level of the noise added to the interpretable features.
    (b) Explanation test error vs. the total number of interpretable features.}
    \vspace{-1ex}
\end{figure}

We train baseline deep architectures (CNN on MNIST and LSTM on IMDB) and their CEN variants.
For MNIST, $\zv$ is either pixels of a scaled down image (\texttt{pxl}) or HOG features (\texttt{hog}).
For IMDB, $\zv$ is either a bag of words (\texttt{bow}) or a topic vector (\texttt{tpc}) produced by a pre-trained topic model.

\textbf{The effect of noisy features.}
In this experiment, we inject noise\footnote{We use Gaussian noise with zero mean and select variance for each signal-to-noise ratio level appropriately.} into the features $\zv$ and ask LIME and CEN to fit explanations to the noisy features.
The predictive performance of the produced explanations on noisy features is given on Fig.~\ref{fig:signal-to-noise}.
Note that after injecting noise, each data point has a noiseless representation $\xv$ and noisy $\tilde\zv$.
Since baselines take only $\xv$ as inputs, their performance stays the same and, regardless of the noise level, LIME ``successfully'' overfits explanations---it is able to almost perfectly approximate the decision boundary of the baselines using very noisy features.
On the other hand, performance of CEN gets worse with the increasing noise level indicating that model fails to learn when the selected interpretable representation is low quality.

\textbf{The effect of feature selection.}
Here, we use the same setup, but instead of injecting noise into $\zv$, we construct $\tilde\zv$ by randomly subsampling a set of dimensions.
Fig.~\ref{fig:incomplete-features} demonstrates the result.
While performance of CENs degrades proportionally to the size of $\tilde\zv$, we see that, again, LIME is able to fit explanations to the decision boundary of the original models despite the loss of information.

These two experiments indicate a major drawback of explaining predictions \emph{post-hoc}: when constructed on poor, noisy, or incomplete features, such explanations can overfit the decision boundary of a predictor and are likely to be misleading.
For example, predictions of a perfectly valid model might end up getting absurd explanations which is unacceptable from the decision support point of view.

\subsection{Explanations as a Regularizer}\label{sec:performance}

In this part, we compare CENs with baselines in terms of performance.
In each task, CENs are trained to simultaneously generate predictions and construct explanations.
Overall, CENs show very competitive performance and are able to approach or surpass baselines in a number of cases, especially on the IMDB data (see Table~\ref{tab:performance}).
This suggests that forcing the model to produce explanations along with predictions does not limit its capacity.

%!TEX root = ../paper.tex

\begin{figure}[h]
    \centering
    \begin{subfigure}[b]{0.45\textwidth}
        \centering
        \includegraphics[width=\textwidth]{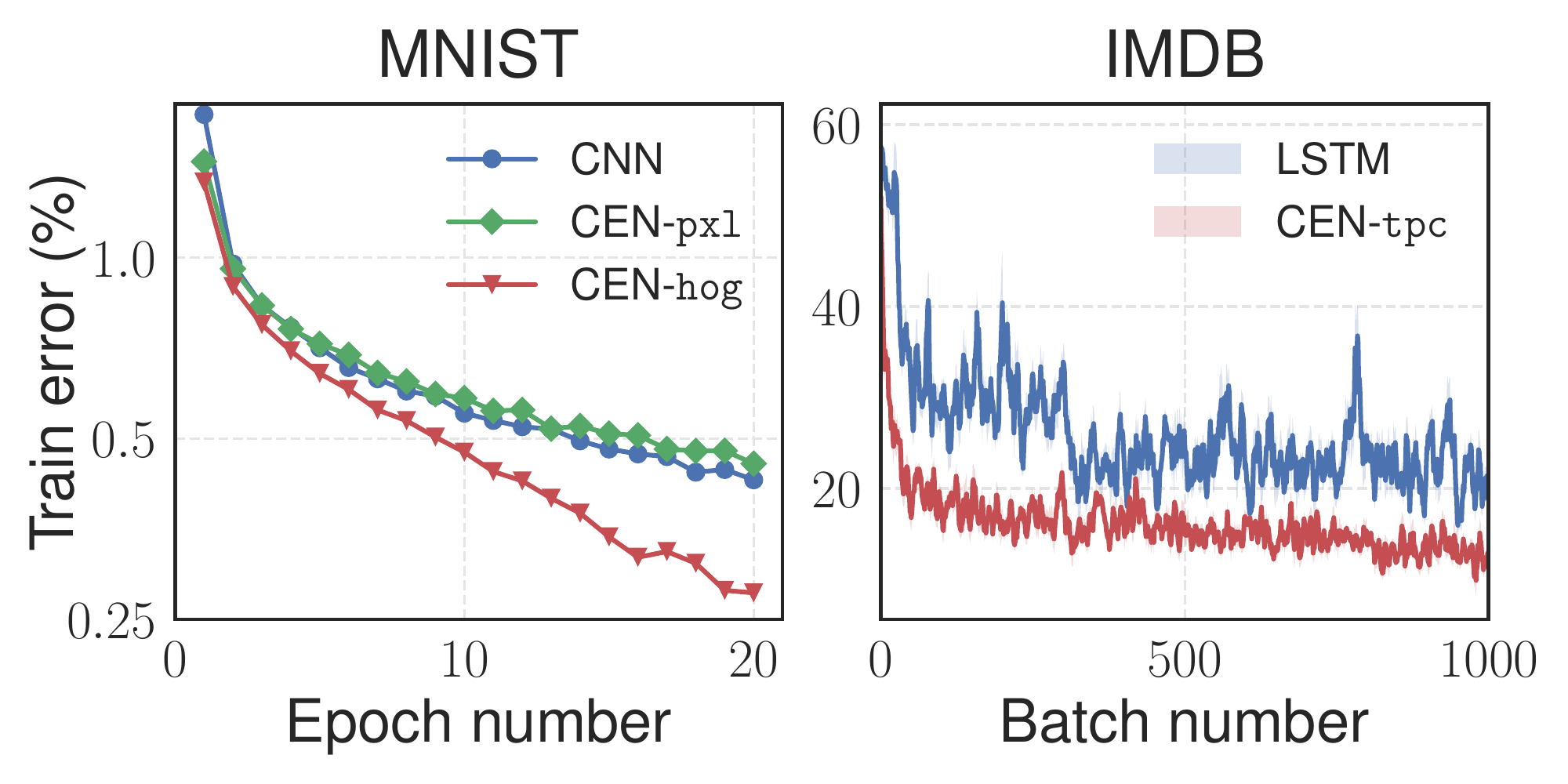}
        \vspace{-4ex}\caption{}\label{fig:convergence}
    \end{subfigure}%
    \begin{subfigure}[b]{0.45\textwidth}
        \centering
        \includegraphics[width=\textwidth]{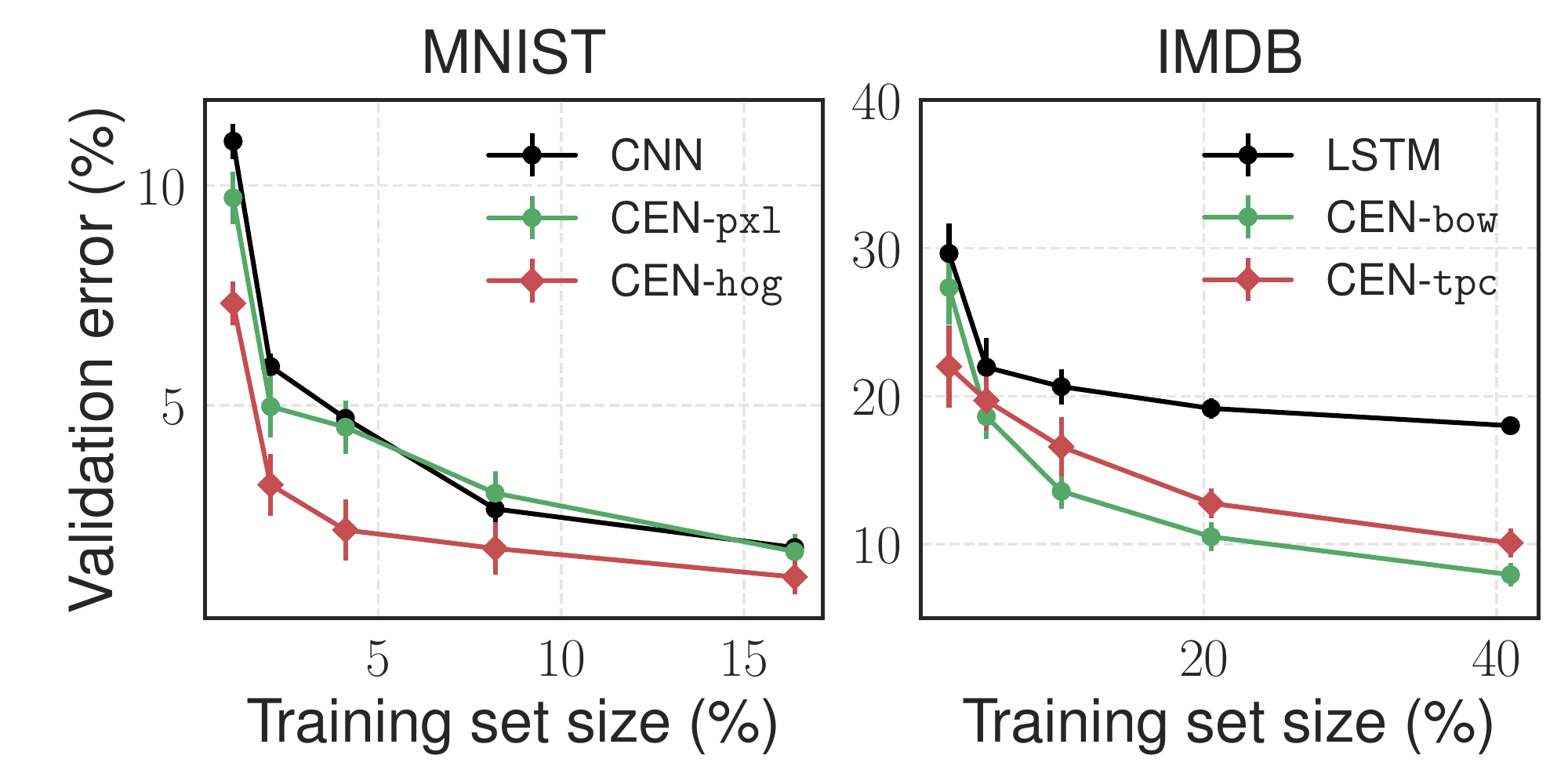}
        \vspace{-4ex}\caption{}\label{fig:sample-complexity}
    \end{subfigure}
    \vspace{-1ex}
    \caption{%
    (a) Training error vs. iteration (epoch or batch) for baselines and CENs.
    (b) Validation error for models trained on random subsets of data of different sizes.}
    \vspace{-1ex}
\end{figure}

Additionally, the ``explanation layer'' in CENs affects the geometry of the optimization problem and causes faster and better convergence (Fig.~\ref{fig:convergence}).
Finally, we train the models on subsets of data (the size varied from 1\% to 20\% for MNIST and from 2\% to 40\% for IMDB) and notice that explanations play the role of a regularizer which strongly improves the sample complexity (Fig.~\ref{fig:sample-complexity}).

%!TEX root = ../paper.tex

\begin{table}[h]
\centering
\caption{%
Performance of the models on classification tasks (averaged over 5 runs; the std. are on the order of the least significant digit).
The subscripts denote the features on which the linear models are built: pixels~(\texttt{pxl}), HOG~(\texttt{hog}), bag-or-words~(\texttt{bow}), topics~(\texttt{tpc}), embeddings~(\texttt{emb}), discrete attributes~(\texttt{att}).%
}\label{tab:performance}
\vspace{1ex}
\small
\def\arraystretch{1.2}
\setlength\tabcolsep{3.2pt}
\begin{tabular}[t]{lr|lr|lrr}
    \toprule
    \multicolumn{2}{c|}{\textbf{MNIST}} &
    \multicolumn{2}{c|}{\textbf{IMDB}} &
    % \multicolumn{2}{c|}{\textbf{20 Newsgroups}} &
    \multicolumn{3}{c}{\textbf{Satellite}} \\
    \midrule
    \textbf{Model}                                  & \textbf{Err (\%)} &
    \textbf{Model}                                  & \textbf{Err (\%)} &
    % \textbf{Model}                                  & \textbf{Err (\%)} &
    \textbf{Model}                                  & \textbf{Acc (\%)} & \textbf{AUC (\%)} \\
    \midrule
    LR$_\texttt{pxl}$                               & $8.00$ &
    LR$_\texttt{bow}$                               & $13.3$ &
    LR$_\texttt{emb}$                               & $62.5$ & $68.1$ \\
    LR$_\texttt{hog}$                               & $2.98$ &
    LR$_\texttt{tpc}$                               & $17.1$ &
    LR$_\texttt{att}$                               & $75.7$ & $82.2$ \\
    CNN                                             & $\mathbf{0.75}$ &
    LSTM                                            & $13.2$ &
    % LSTM                                            & $18.7$ &
    MLP                                             & $77.4$ & $78.7$ \\
    \midrule
    MoE$_\texttt{pxl}$                              & $1.23$ &
    MoE$_\texttt{bow}$                              & $13.9$ &
    MoE                                             & $77.9$ & $\mathbf{85.4}$ \\
    MoE$_\texttt{hog}$                              & $1.10$ &
    MoE$_\texttt{tpc}$                              & $12.2$ &
    CEN                                             & $81.5$ & $84.2$ \\
    CEN$_\texttt{pxl}$                              & $\mathbf{0.76}$ &
    CEN$_\texttt{bow}$                              & $^\star\mathbf{6.9}$ &
    \multirow{2}{*}{VCEN}                           & \multirow{2}{*}{$\mathbf{83.4}$} & \multirow{2}{*}{$84.6$} \\
    CEN$_\texttt{hog}$                              & $\mathbf{0.73}$ &
    CEN$_\texttt{tpc}$                              & $^\star7.8$ &
                                                    &  & \\
    \bottomrule
\end{tabular}\\
{\scriptsize$^\star$Best previous results for similar LSTMs: $8.1\%$ (supervised) and $6.6\%$ (semi-supervised)~\cite{johnson2016supervised}.}
\end{table}

\subsection{Visualizing Explanations}\label{sec:insights}

Finally, we showcase the insights one can get from explanations produced along with predictions.
Particularly, we consider the problem of poverty prediction for household clusters in a Uganda from satellite imagery and survey data.
The $\xv$ representation of each household cluster is a collection of $400 \times 400$ satellite images; $\zv$ is represented by a vector of 65 categorical features from living standards measurement survey (LSMS).
The goal is binary classification of households in Uganda into poor and not poor.
In our methodology, we closely follow the original study of~\citet{jean2016combining} and use a pretrained \VGGF network for embedding the images into a 4096-dimensional space on top of which we build our contextual models.
Note that this datasets is fairly small (642 points), and hence we keep the \VGGF frozen to avoid overfitting.
We note that quantitatively, by conditioning on the VGG features of the satellite imagery, CENs are able to significantly improve upon the sparse linear models on the survey features only (known as the gold standard in remote sensing techniques).

After training CEN with a dictionary of size 32, we discover that the encoder tends to sharply select one of the two explanations (M1 and M2) for different household clusters in Uganda (see Fig.~\ref{fig:satellite-models} and also Fig.~\ref{fig:satellite-models-full} in appendix).
In the survey data, each household cluster is marked as either urban or rural; we notice that, conditional on a satellite image, CEN tends to pick M1 for urban areas and M2 for rural (Fig.~\ref{fig:satellite-barplots}).
Notice that explanations weigh different categorical features, such as reliability of the water source or the proportion of houses with walls made of unburnt brick, quite differently.
When visualized on the map, we see that CEN selects M1 more frequently around the major city areas, which also correlates with high nightlight intensity in those areas (Fig.~\ref{fig:satellite-map-cen},\ref{fig:satellite-map-nl}).
High performance of the model makes us confident in the produced explanations (contrary to LIME as discussed in Sec.~\ref{sec:consistency}) and allows us to draw conclusions about what causes the model to classify certain households in different neighborhoods as poor.

%!TEX root = ../paper.tex

\begin{figure}[b]
\begin{subfigure}[b]{0.17\textwidth}
    \centering
    \includegraphics[width=\textwidth]{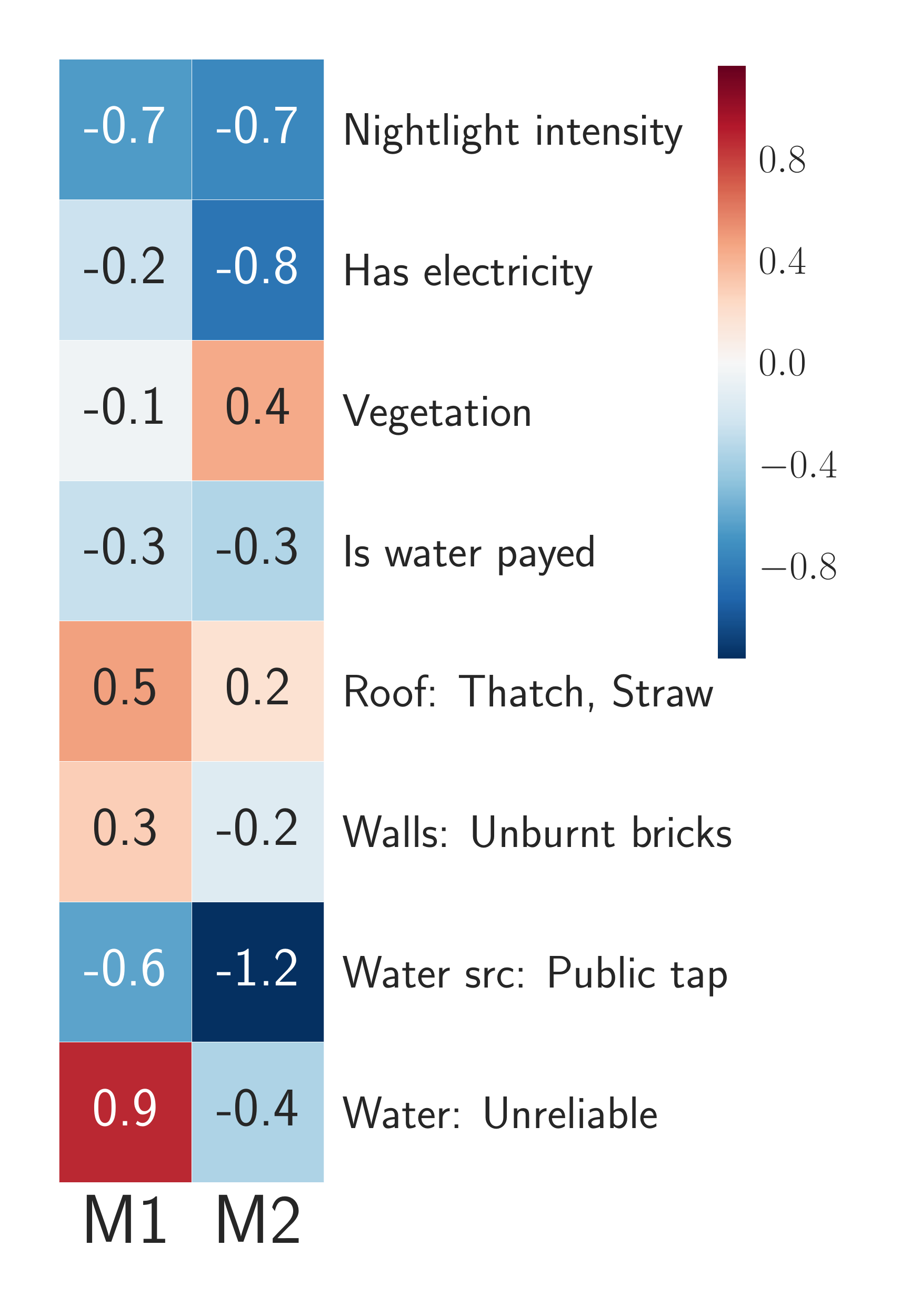}%
    \vspace{-2ex}\caption{}\label{fig:satellite-models}
\end{subfigure}
\hfill
\begin{subfigure}[b]{0.17\textwidth}
    \centering
    \includegraphics[width=\textwidth]{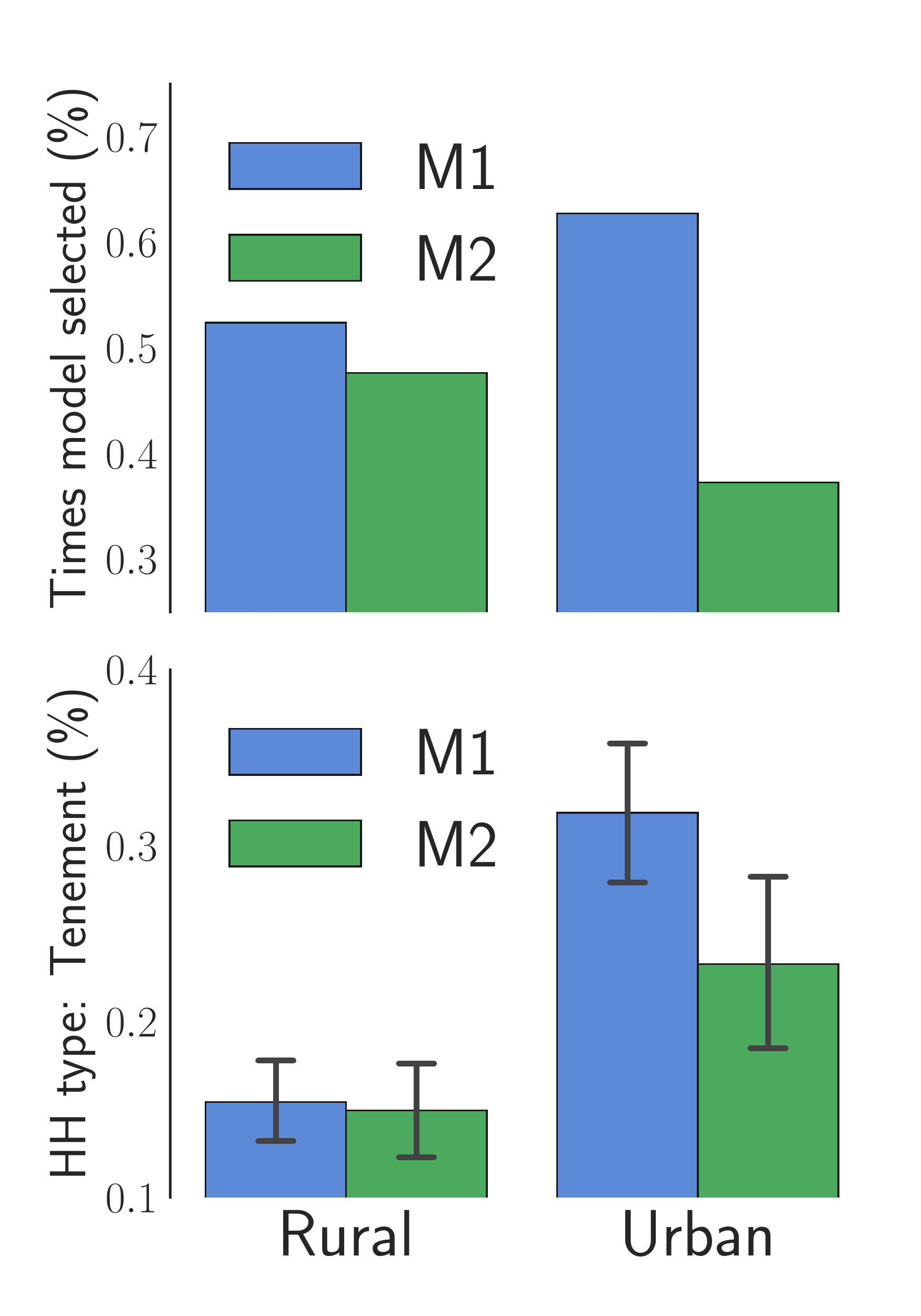}%
    \vspace{-2ex}\caption{}\label{fig:satellite-barplots}
\end{subfigure}
\hfill
\begin{subfigure}[b]{0.30\textwidth}
    \centering
    \includegraphics[width=\textwidth]{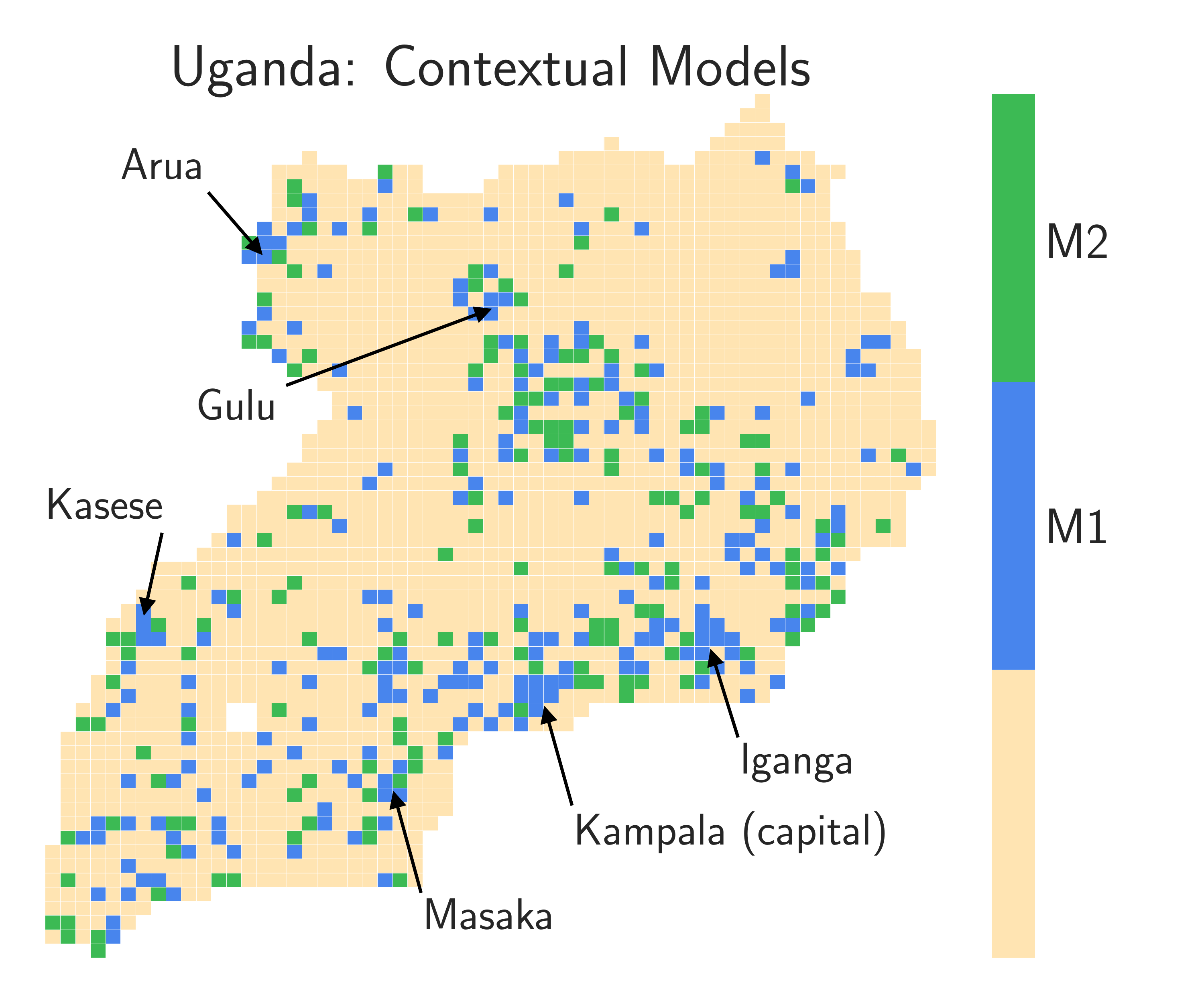}%
    \vspace{-2ex}\caption{}\label{fig:satellite-map-cen}
\end{subfigure}
\hfill
\begin{subfigure}[b]{0.30\textwidth}
    \centering
    \includegraphics[width=\textwidth]{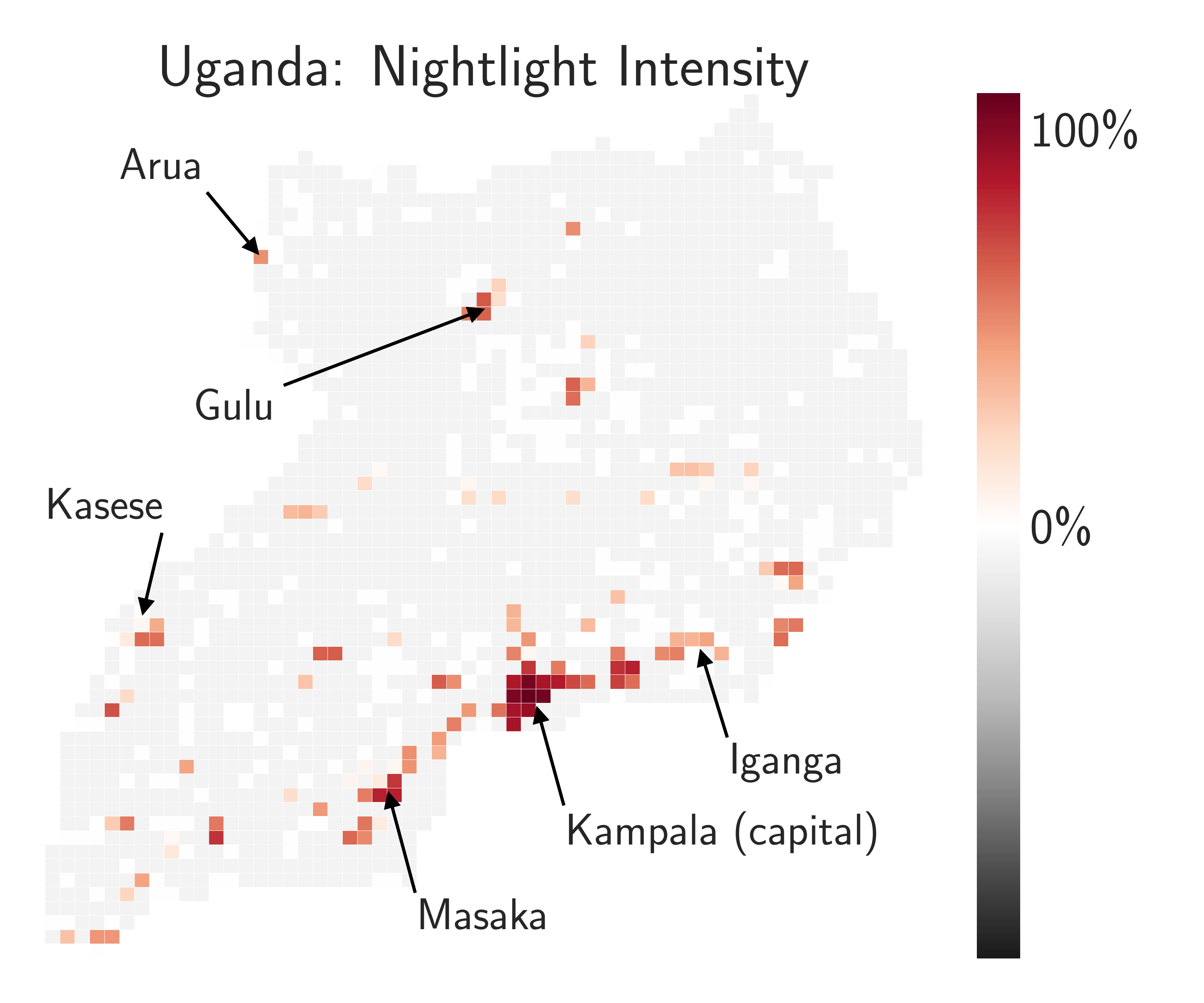}%
    \vspace{-2ex}\caption{}\label{fig:satellite-map-nl}
\end{subfigure}
\caption{%
Qualitative results for the Satellite dataset:
(a) Weights given to a subset of features by the two models (M1 and M2) discovered by CEN.
(b) How frequently M1 and M2 are selected for areas marked rural or urban (top) and the average proportion of Tenement-type households in an urban/rural area for which M1 or M2 was selected.
(c) M1 and M2 models selected for different areas on the Uganda map.
M1 tends to be selected for more urbanized areas while M2 is picked for the rest.
(d) Nightlight intensity of different areas of Uganda.}
\label{fig:satellite}
\end{figure}

\clearpage
\bibliography{references}
\bibliographystyle{unsrtnat}

\clearpage
\appendix
%!TEX root = ../paper.tex

\section{Appendix}\label{sec:appendix}

%!TEX root = ../paper.tex

\begin{figure}[h]
\begin{subfigure}[b]{\textwidth}
    \centering
    \includegraphics[width=\textwidth]{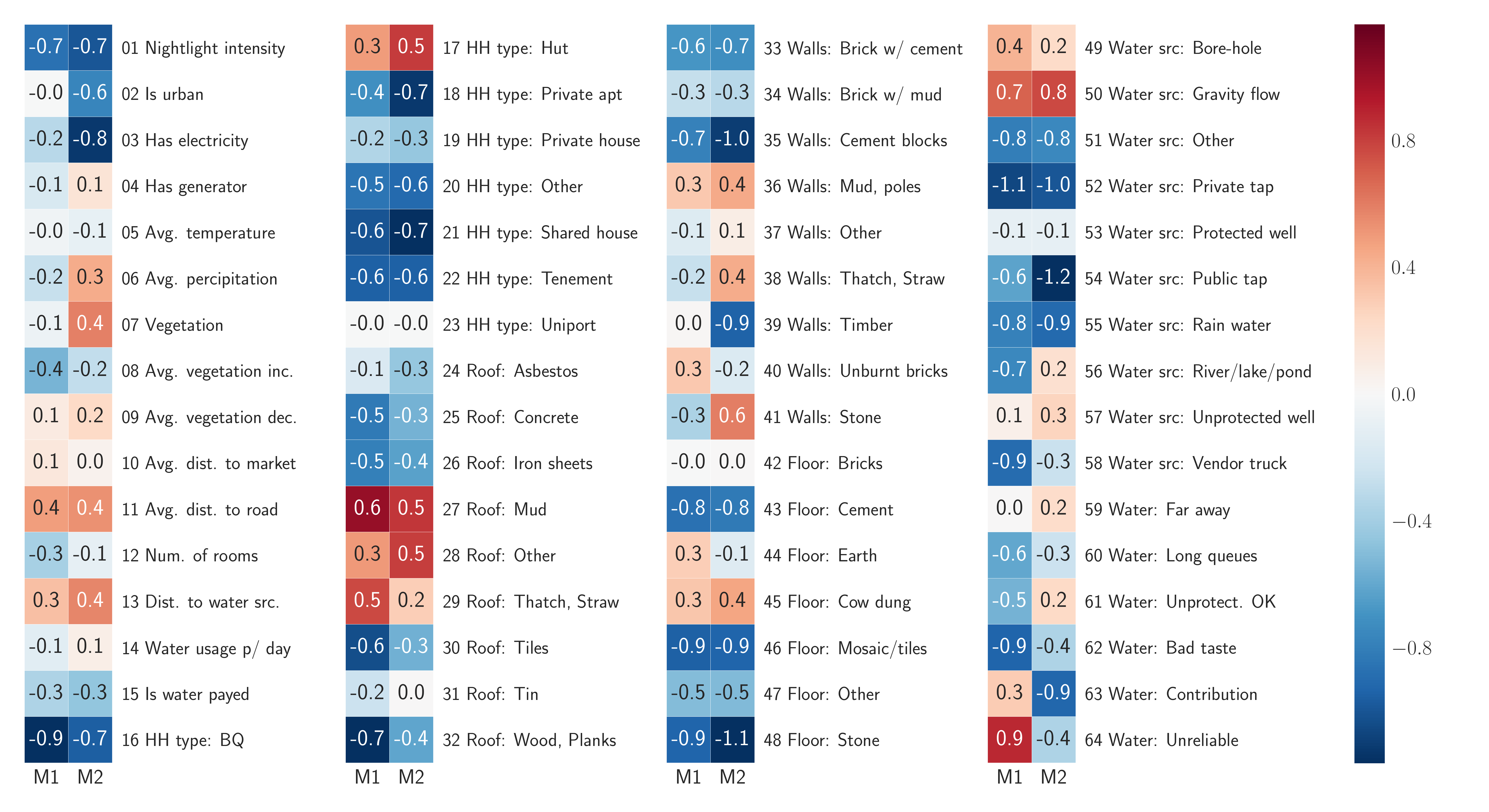}
    \vspace{-4ex}\caption{%
        Full visualization of explanations M1 and M2 learned by CEN on the poverty prediction task.
    }\label{fig:satellite-models-full}
\end{subfigure}
\begin{subfigure}[b]{\textwidth}
    \centering
    \includegraphics[width=\textwidth]{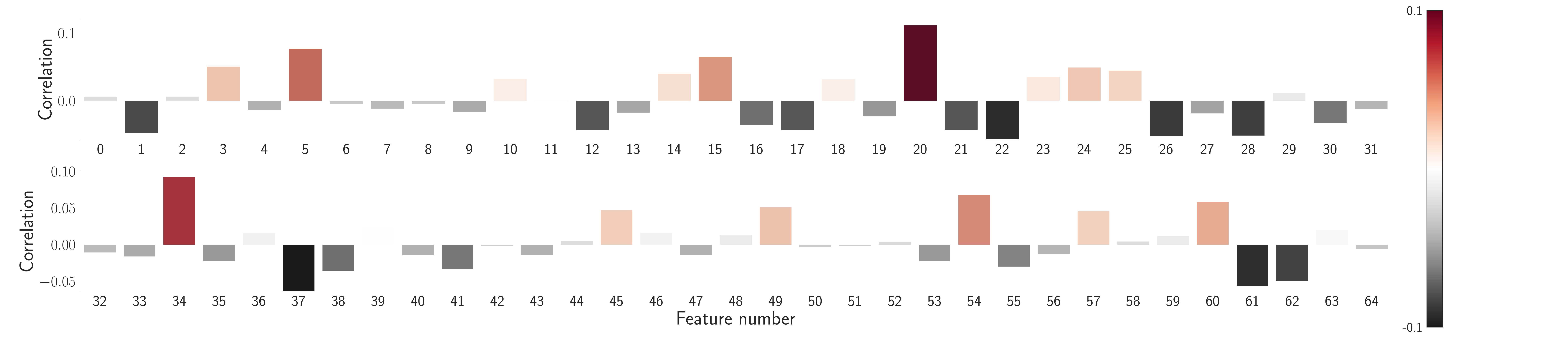}
    \vspace{-4ex}\caption{%
        Correlation between the selected explanation and the value of a particular survey variable.
    }\label{fig:satellite-models-feature-corr}
\end{subfigure}
\caption{%
Additional visualizations for the poverty prediction task.}
\label{fig:satellite-appendix}
\end{figure}

\section{Details on Consistency of Explanations}\label{app:consistency}

We provide a detailed description of the experimental setup used for our analysis in Section~\ref{sec:consistency}.

\end{document}